\documentclass{bmvc2k}

\title{You said that?}

\addauthor{Joon Son Chung$^\star$}{http://www.robots.ox.ac.uk/~joon}{1}
\addauthor{Amir Jamaludin$^\star$}{http://www.robots.ox.ac.uk/~amirj}{1}
\addauthor{Andrew Zisserman}{http://www.robots.ox.ac.uk/~az}{1}

\addinstitution{ 
 Visual Geometry Group\\
Department of Engineering Science\\
 University of Oxford\\ 
 Oxford, UK
}

\runninghead{Chung, Jamaludin, Zisserman}{You said that?}

\def\eg{\emph{e.g}\bmvaOneDot}

\def\etal{\emph{et al}\bmvaOneDot}

\def\subsub{\vspace{10pt}}

\newcommand\blfootnote[1]{%
  \begingroup
  \renewcommand\thefootnote{}\footnote{#1}%
  \addtocounter{footnote}{-1}%
  \endgroup
}

\begin{document}

\maketitle
\blfootnote{\hspace{-0.5cm}$^\star$ These authors contributed equally to this work.}

\begin{abstract}
We present a method for generating a video of a talking face.
The method takes as inputs: (i) still images
of the target face, and (ii) an audio speech segment; and outputs a video
of the target face lip synched with the audio. 
The method runs in real time
and is applicable to faces and audio not seen at training time.

To achieve this we propose an
encoder-decoder CNN model that uses a joint embedding of the face and
audio to generate synthesised talking face video frames. The model is
trained on tens of hours of unlabelled videos.

We also show results of re-dubbing videos using speech from a
different person.

\end{abstract}


\begin{figure}[h]
	\centering
	\fbox{\includegraphics[width=0.95\textwidth]{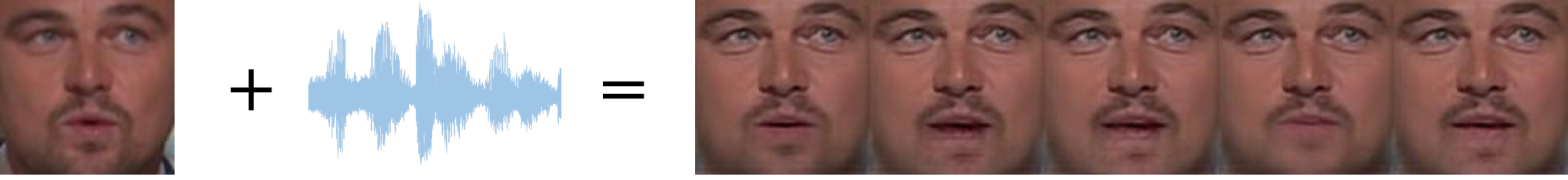}}
	\caption{The Speech2Vid model generates a video of a talking face, given 
	still images of the person and a speech segment. The model 
takes an image of the target face and an audio segment, and outputs a video
of the target face lip synched with the audio. Note that the target face need not be in the  training dataset {\em i.e.}\ the Speech2Vid is applicable to unseen images and speech.}
	\label{fig:teaser}
\end{figure}

\section{Introduction}
\label{sec:intro}
There has been much work recently in the area of transforming one 
modality to another. Image to text is the most prominent, \eg in
caption generation~\cite{Vinyals15,karpathy2015deep,Xu15}, but there is also
video to sound \eg 
\cite{OwensIMTAF16}, text to image \cite{ReedAYLSL16}, or in fact a
mixture of different mediums \eg video and audio to text
\cite{Chung17}. This paper considers the case of 
audio to video.

We propose a method to generate videos of a talking face using only an
audio speech segment and a face image of the target identity (audio
and image to video). The speech segment need not be spoken originally
by the target person (see Figure~\ref{fig:teaser}). We dub the
approach {\em Speech2Vid}.  Our method differs from previous 
approaches for this task (see related work below) in that
instead of learning phoneme to viseme mappings, we learn the correspondences between raw audio and video data directly. 
By focusing on the speech portion of audio and tight facial regions of
speakers in videos, the Speech2Vid model is able to produce natural-looking
videos of a talking face at test time even when using an image and 
audio outside of the
training dataset.

The key idea of the approach is to learn a joint embedding of the
target face and speech segment that can be used to generate a frame of
that face saying (lip synched with) the speech segment. Thus the inputs
are still images of the face (that provides the identity, but is not
speaking the target segment) and the target speech segment; and the
generated output is the target face speaking the segment. The Speech2Vid model
can be learnt from unlabelled videos, as shown in Figure~\ref{fig:samplingIdentity}.

In the following, we first describe the 
automatic pipeline  to 
prepare the video dataset used to train the generation network (Section~\ref{sec:dataset}).
The architecture and training of the Speech2Vid model is given in 
Section~\ref{sec:s2v}. Finally, Section~\ref{sec:exp}, assesses variations on the architecture, including using multiple images of the identity as input, and shows an application of the model to re-dubbing videos
by visually blending  the generated face into the source video frame.

\subsection{Related Work}
\label{subsec:rWorks}
There are various works that proposed methods to generate or synthesise videos of talking heads from either audio or text sources. Fan~\etal~\cite{fan2015photo} introduced a method to restitch the lower half of the face via a bi-directional LSTM to re-dub a target video from a different audio source. The LSTM selects a target mouth region from a dictionary of saved target frames, rather than generating the image, so it requires a sizeable amount of video frames of the unique target identity to choose from.
Similarly, Charles~\etal~\cite{charles2016virtual} uses phonetic labels to select frames from a dictionary of mouth images.
Wan~\etal~\cite{wan2013photo} proposed a method to synthesise a talking head via an active appearance model with the ability to control the emotion of the talking avatar, but they are constrained to the unique model trained by the system.
Garrido~\etal~\cite{garrido2015vdub} synthesises talking faces on target speakers by transferring the mouth shapes from the video of the dubber to the target video, but this method requires the video footage of the dubber's mouth saying the speech segment, whereas our method learns the relationship between the sound and the mouth shapes.

Our training approach is based on unsupervised learning, in our case
from tens of hours of people talking. One of the earliest example of
unsupervised learning of the representation of data using neural
networks is the autoencoder by Hinton and
Salakhutdinov~\cite{HintonSalakhutdinov2006b}. Further improvements in
representation learning, for example the variational auto-encoders by
Kingma~\etal~\cite{KingmaW13}, opens up the possibility of generating
images using neural networks. Moving forward, current research shows
adversarial training proposed by~\cite{GoodfellowPMXWOCB14} works well
for generating natural-looking images; conditional
generative models~\cite{van2016conditional} are able to generate images based on
auxilary information such as a class label. Our Speech2Vid model is
closest in spirit to the image-to-image model by
Isola~\etal~\cite{pix2pix2016} in that we generate an output that
closely resembles the input, but in our case we have both audio and
image data as inputs.



\section{Dataset}
\label{sec:dataset}

This section describes our multi-stage strategy to prepare
a large-scale
dataset to train the generation network. 
We obtain tens of hours of visual face sequences aligned with
spoken audio.

\begin{figure}[h]
	\centering
	\includegraphics[width=0.9\textwidth]{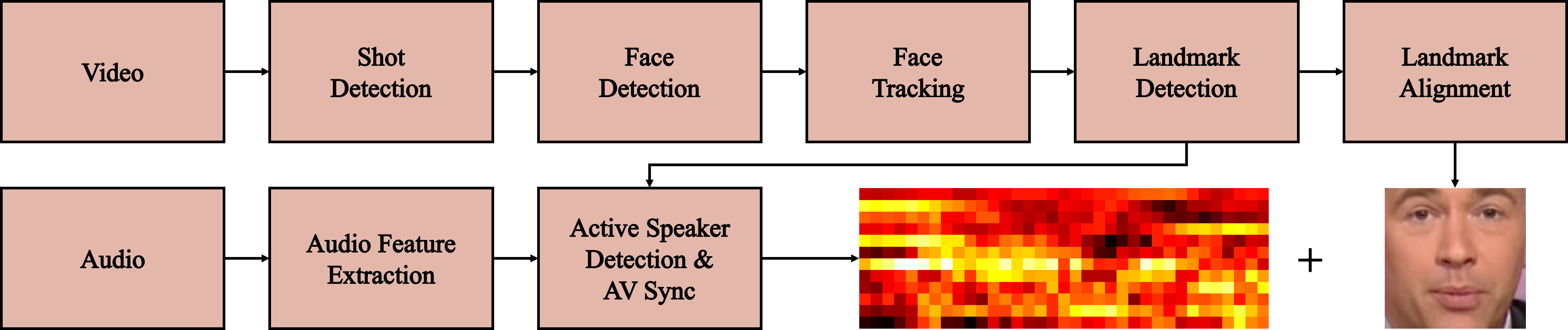}
	\caption{Data preparation pipeline.}
	\label{fig:prepPipeline}
\end{figure}

The principal stages are:
(i) detect and track all face appearances in the video;
(ii) determine who is speaking in the video;
and (iii) align the detected face image to the canonical face.
The pipeline is summarised in Figure~\ref{fig:prepPipeline},
and the details are discussed in the following paragraphs. 

\subsub\noindent\textbf{Video description.}
We train the Speech2Vid model on videos
from the VoxCeleb~\cite{Nagrani17}
and LRW~\cite{Chung16} datasets.
These datasets consist of celebrity interviews and
broadcast news, which provide
ideal training data for this task, given that a large
proportion of the face tracks are front-facing, and of
high quality. 
Moreover, the words are generally clearly spoken without
too much background noise, and hence provide an easier
learning environment for the network. 

\subsub\noindent\textbf{Face tracking.}
The face tracking pipeline is based on~\cite{Chung16a}.
First, the shot boundaries are determined by comparing
colour histograms~\cite{Lienhart01}
to find the within-shot
frames for which tracking is to be run. 
The HOG-based DLIB face detector~\cite{King09} is used to detect
face appearances on every frame of the video.
The face detections are grouped into face tracks using a 
KLT detector~\cite{Lucas81}.
Facial landmarks are extracted 
using the regression-tree based
method of~\cite{Kazemi14}. 

\subsub\noindent\textbf{Active speaker detection and AV synchronisation.} 
SyncNet~\cite{Chung16a} provides a joint embedding of the
audio and visual face sequences in a video, which can be
used to determine who is speaking in a multi-speaker video scene.
Moreover, the same method is used to correct the lip-sync
error in the broadcast video, 
which can be crucial for precisely locating the corresponding
mouth image for the audio sample.  

\begin{figure}[h]
	\centering
	\includegraphics[width=0.19\textwidth]{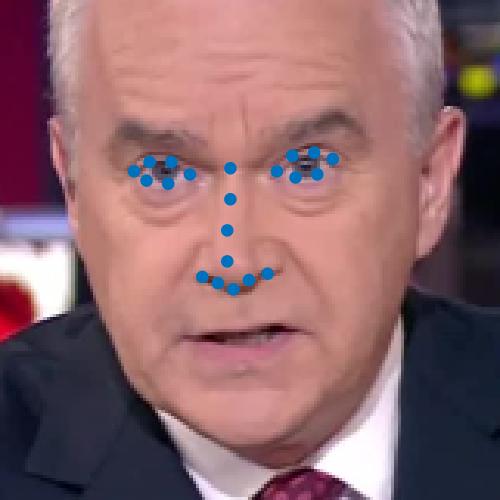}
	\includegraphics[width=0.19\textwidth]{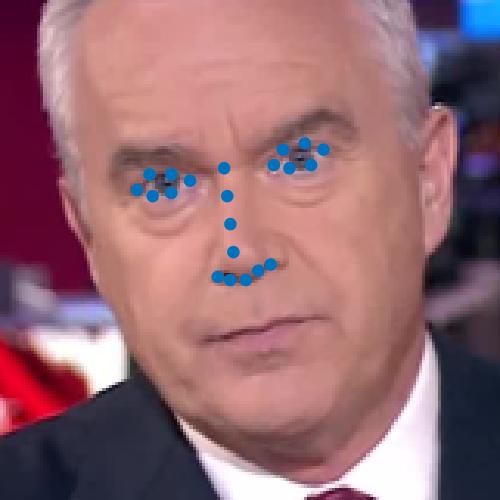}
	\includegraphics[width=0.19\textwidth]{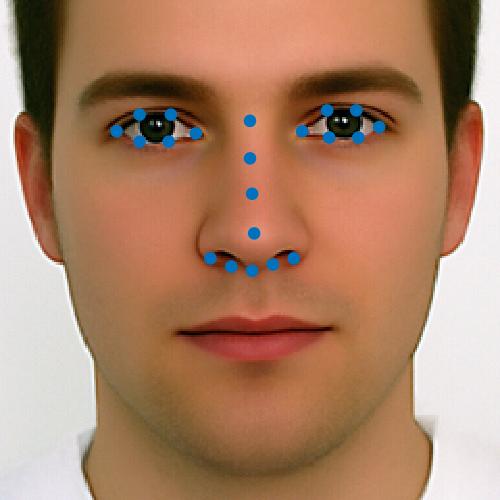}
	\includegraphics[width=0.19\textwidth]{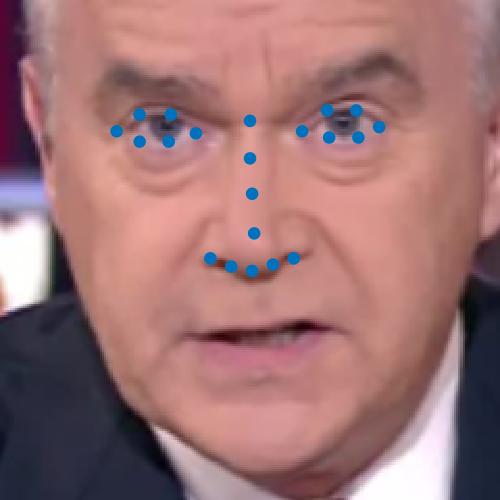}
	\includegraphics[width=0.19\textwidth]{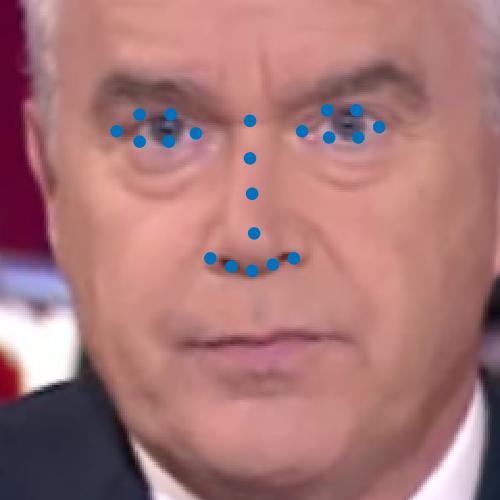}
	\caption{
	{\bf Left pair:} 	Face images before registration;
	{\bf Middle:} 		Canonical face;
	{\bf Right pair:} 	Face images after registration with the canonical face.
	}
	\label{fig:align}
\end{figure}

\subsub\noindent\textbf{Spatial registration.}
In order to establish spatial correspondance between the
input face (that provides the identity to the encoder) and the output
face (from the decoder) in training from 
the ground truth frames, 
we register the facial landmarks between the two images.
This is done by performing a similarity transformation 
(scale, rotation and translation) between 
the faces and an exemplar face with canonical position 
(Figure~\ref{fig:align} middle).
Only the landmarks on the eyes and the nose, not the mouth, are used to
align the face image, as the mouth movements contain 
the information that we wish to capture. 

\subsub\noindent\textbf{Data statistics.}
The train-validation split is given
in the VoxCeleb and LRW datasets.
For every valid face track, we extract
every 5th frame and the corresponding
audio as samples for training and 
validation.
Statistics on the dataset is given in 
Table~\ref{table:datastat}.

\begin{table}[th!]
\centering

\begin{tabular}{| l | r | r | }
  \hline \textbf{Set} & \textbf{\# Hours} & \textbf{\# Samples}  \\ 
  \hline \hline 

    Train & 37.7 & 678,389 \\ \hline 
    Val   & 0.5 & 9,287 \\ \hline 
\end{tabular} 

\caption{Dataset statistics}
\label{table:datastat}

\end{table}


\section{The Speech2Vid Model}
\label{sec:s2v}

\begin{figure}[h]
	\centering
	\includegraphics[width=0.85\textwidth]{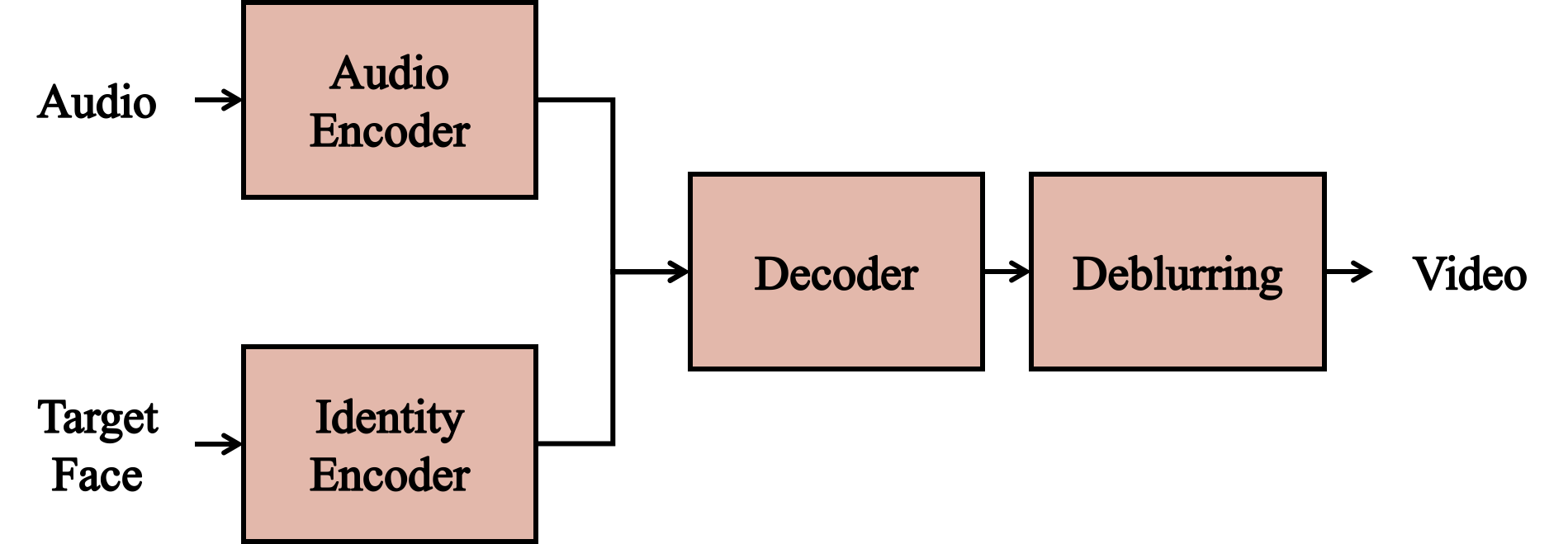}
	\caption{The overall Speech2Vid model is a combination of two encoders taking in two different streams of data, audio and image, a decoder that generates images corresponding to the audio, and a CNN deblurring module that refine the output frames.}
	\label{fig:modelPipe}
\end{figure}

Our main goal at test time is to generate a video of a talking face
given two inputs: (i) an audio segment, and (ii) still images of the
target identity (frontal headshot). The Speech2Vid model (summarised
in Figure~\ref{fig:modelPipe} at the block level), consists of four main components: an
audio encoder, an identity image encoder, a talking face image
decoder, and a deblurring module. For a given input sample, the model
generates one frame of image output that best represents the audio
sample at a specific time step. The model generates the video on a
frame-by-frame basis by sliding a 0.35-second
window over the audio sequence.

\subsection{Input Representations}
\label{subsec:iRepresentations}

This section describes the input representations for the audio and identity. These inputs are fed into separate modules in the network in the forms of 0.35-second audio and a still image of the target identity.

\begin{figure}[h]
	\centering
	\includegraphics[width=0.95\textwidth]{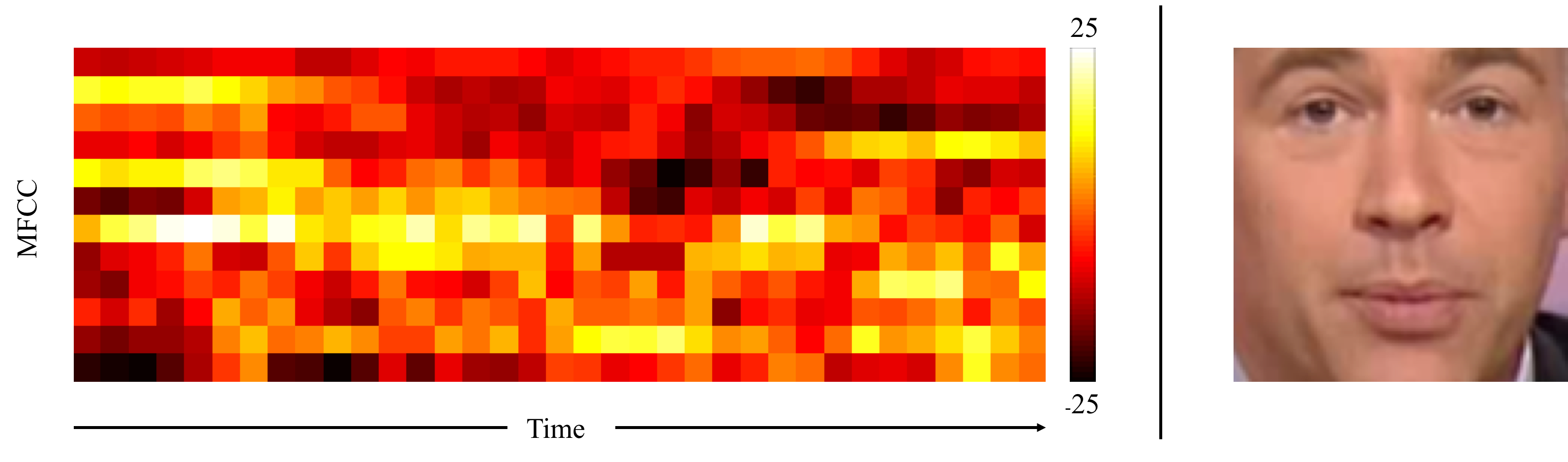}
	\caption{Inputs to the Vid2Speech model. \textbf{Left:} MFCC heatmap for the 0.35-second time period. The 12 rows in the matrix  represent the power of the audio at different frequencies. \textbf{Right:} Still image of the speaker.}
	\label{fig:temporal}
\end{figure}

\subsub\noindent\textbf{Audio.}
The input to the audio encoder are Mel-frequency cepstral coefficients (MFCC) values extracted from the raw audio data. The MFCC values are made up of individual coefficient each representing a specific frequency band of the audio short-term power on a non-linear mel scale of frequency; 13 coefficients are calculated per sample but only the last 12 are used in our case. Each sample fed into the audio encoder is made up of 0.35-second input audio data with a sampling rate of 100Hz resulting in 35 time steps. Each encoded sample can be viewed as a $12 \times 35$ heatmap where each column represents MFCC features at each time step (see Figure~\ref{fig:temporal}). 

\subsub\noindent\textbf{Identity.} The input to the identitiy encoder is a single still  image 
with dimensions $112 \times 112 \times 3$. In Section~\ref{subsec:multieg}, we also experiment with having multiple still images as the input to the identity encoder instead of one, which significantly improves the output video quality.

\subsection{The Architecture}
\label{subsec:arch}

\begin{figure}[h]
	\centering
	\includegraphics[width=1.0\textwidth]{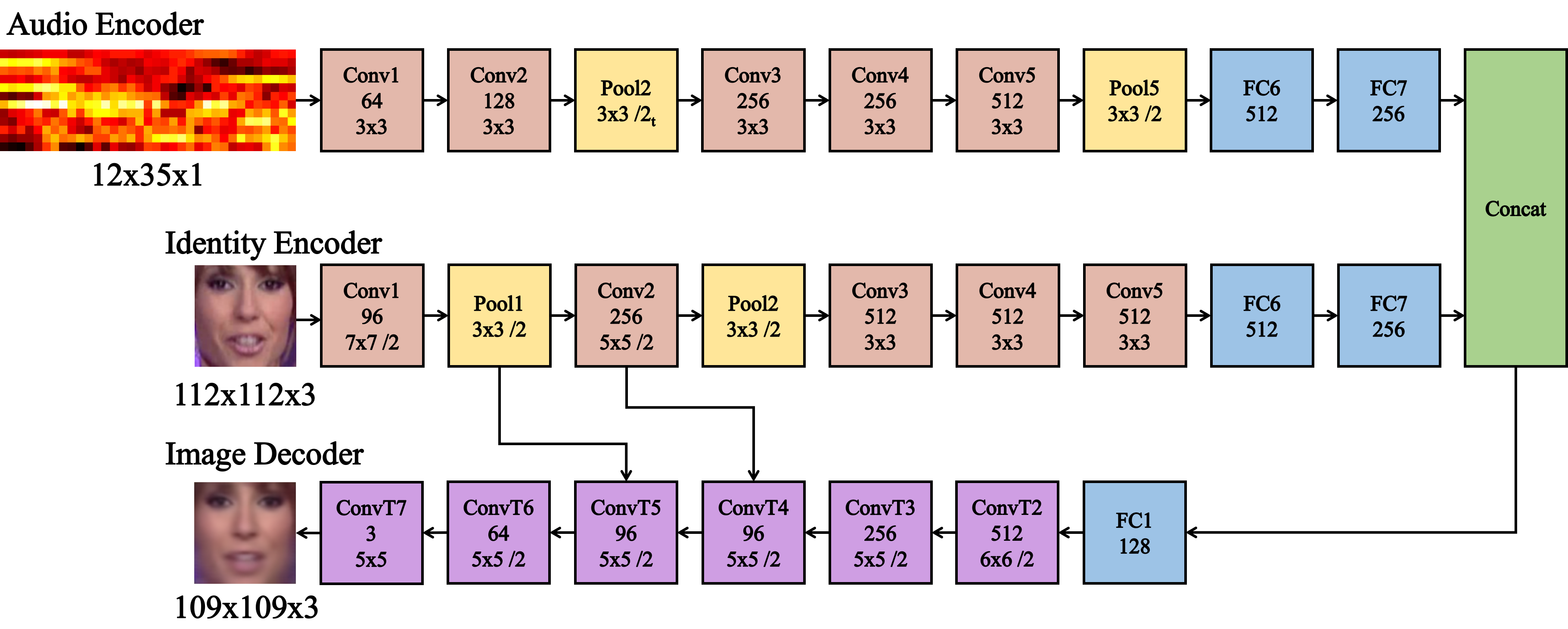}
	\caption{The three modules in the Speech2Vid model. From top to bottom: (i) audio encoder, (ii) identity encoder with a single still image input, and (iii) image decoder. $/2$ refers to the stride of each kernel in a specific  layer which is normally of equal stride in both spatial dimensions except for the Pool2 layer in which we use stride 2 in the timestep dimension (denoted by $/2_t$). The network includes two skip connections between the identity encoder and  the image decoder.}
	\label{fig:encoderdecoder}
\end{figure}

The Speech2Vid architecture is given in
Figure~\ref{fig:encoderdecoder}. We describe the
three modules (audio encoder, the identity encoder, and the image decoder)
in the
following paragraphs. Note, these three modules 
are trained together. The deblurring module (described below in Section~\ref{subsec:deblur}) is trained separately. 

\subsub\noindent\textbf{Audio encoder.}
We use a convolutional neural network originally designed for
image recognition. The layer configurations
is based on AlexNet~\cite{Krizhevsky12} and VGG-M~\cite{Chatfield14}, 
but filter sizes are adapted for the unusual input dimensions.
This is similar to the configuration used to learn audio embedding in~\cite{Chung16a}. 

\subsub\noindent\textbf{Identity encoder.} 
Ideally, the identity vector produced by the encoder should have features unique for facial recognition and as such we use a VGG-M network pre-trained on the VGG Face dataset~\cite{Parkhi15}. The dataset includes 2.6M images of 2.6K unique identities. Only the weights of the convolutional layers are used in the encoder, while the weights of the fully-connected layers are reinitialized. 

\subsub\noindent\textbf{Image decoder.} 
The decoder takes as input the concatenated feature vectors of the FC7 layers of the audio and identity encoders (both 256-dimensional). The features vector is gradually upsampled, layer-by-layer via transposed convolutions. See details in Figure~\ref{fig:encoderdecoder}. The network features two skip connections to help preserve the defining features of the target identity -- this is done by concatenating the encoder activations with the decoder activations (as suggested in~\cite{ronneberger2015u}) at the locations shown in the network diagram.

\subsub\noindent\textbf{Loss function.}
An $L_1$ loss is used (Equation~\ref{eqn:loss}), rather than $L_2$ that is more commonly used for image generation and in auto-encoders, as $L_1$ tends to encourage less blurring~\cite{pix2pix2016}.
\begin{equation}
	\mathcal{L}_= \sum_{n=1}^{N}||\hat{y}_n - y_n||
	\label{eqn:loss}
\end{equation}

\begin{figure}[h]
	\centering
	\includegraphics[width=1.0\textwidth]{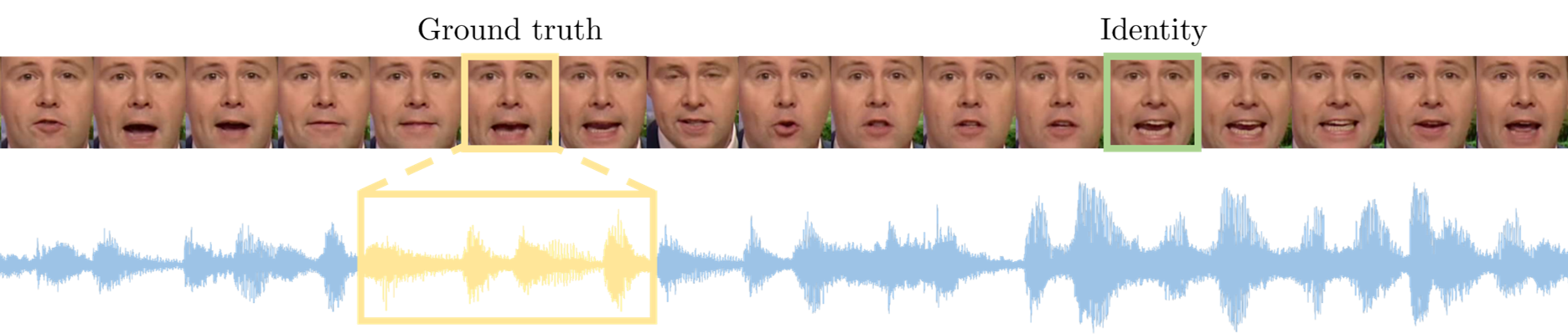}
	\caption{Sampling strategy for identity images during training. Identities are randomly sampled from future frames far from actual audio/output image samples.}
	\label{fig:samplingIdentity}
\end{figure}

\subsub\noindent\textbf{Training protocol.}
The network is trained on the video dataset described in
Section~\ref{sec:dataset}.  During training, the ground truth output
image of the target identity speaking the audio segment is used as
supervision. The image is taken from the middle frame of the video in
the 0.35-second sampling window.
The image for the input
identity 
of the speaker is randomly sampled from a different point in time, as shown in 
Figure~\ref{fig:samplingIdentity}). 
When multiple still images are used for the input identity (Section~\ref{subsec:multieg})
we randomly sample multiple images from the same video stream.

\subsub\noindent\textbf{Discussion.}
The network architecture is based purely on ConvNets,
as opposed to the recurrent architectures often used for
tasks relating to time sequences.
The rationale is that the mouth shape of the speaker
does not depend on anything other than the phoneme
that is being said at the exact moment, and the 
long term context is unimportant. We find that the 0.35-second
window is more than enough to capture this information. At test time, the video is generated frame-by-frame by sliding a temporal window across the entire audio segment while using the same single identity image.

\subsub\noindent\textbf{Implementation details.} Our implementation is based on the MATLAB toolbox MatConvNet \cite{Vedaldi15}
and trained on a NVIDIA Titan X GPU with 12GB memory. The network is
trained with batch normalisation and a fixed learning rate of $10^{-5}$ using stochastic gradient descent with momentum.
The training was stopped after 20 epochs, or when the performance on the validation set
stops improving, whichever is sooner. 

At test time, the network (including the deblurring layers) runs faster than twice real-time on a GPU.
This can be further accelerated by pre-computing and saving the features from the identity encoder module,
rather than running this for every frame.
In the case of redubbing video, the 
the output video is generated at the same frame rate as the original video.

\subsection{Deblurring module} 
\label{subsec:deblur}

\begin{figure}[h]
	\centering
	\includegraphics[width=1.0\textwidth]{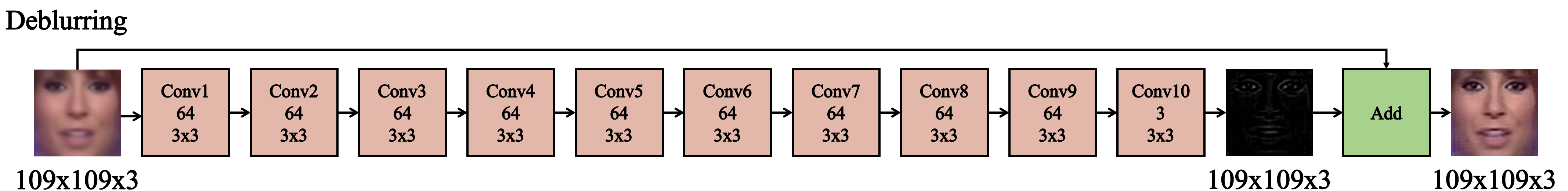}
	\caption{Deblurring CNN module}
	\label{fig:deblurring}
\end{figure}

CNNs trained to generate images with $L_1$ and $L_2$ losses
tend to produce blurry images~\cite{pathak2016context,zhang2016colorful}.
To mitigate this problem, 
we train a separate deblurring CNN to sharpen the images produced by
the Speech2Vid model.
The model is inspired by VDSR~\cite{Kim_2016_VDSR},
which uses a residual connection between the input and 
output, so that the network only has to learn the image difference.
Our implementation has 10 convolutional and ReLU layers, and
the layer configuration 
is shown in Figure~\ref{fig:deblurring}.

\begin{figure}[h]
	\centering
	\includegraphics[width=0.24\textwidth]{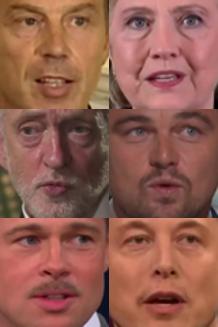}~~~~~~
	\includegraphics[width=0.24\textwidth]{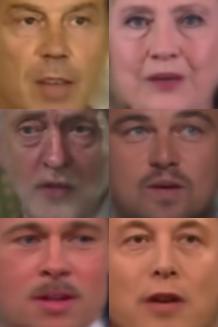}~~~~~~
	\includegraphics[width=0.24\textwidth]{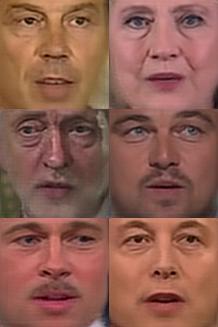}~
	
	\caption{
	Deblurring CNN input and output. 
	{\bf Left:}		  	Original face image (ground truth);
	{\bf Middle:}		Input to the deblurring CNN;
	{\bf Right:} 	 	Restored face image using the deblurring CNN.
	}
	\label{fig:deblur_data}
\end{figure}

We train the network on artificially blurred face images
(Figure~\ref{fig:deblur_data}),
as opposed to training the network end-to-end together
with the generator network.
This is because the alignments between the input image,
the target (ground truth) image and the
generated image are not perfect even after the spatial 
registration (of Section~\ref{sec:dataset}),
and thus avoid the deblurring network having to learn the residual
coming from the misalignment.

The images that we ask the CNN to deblur are relatively
homogeneous in content (they are all face images),
and we find that the CNN performs very well in
sharpening the images under this constraint.


\section{Experiments}
\label{sec:exp}
{\bf The results are best seen in video format.
Please refer to the online examples.}
\begin{figure}[h]
	\centering
	\includegraphics[width=0.7\textwidth]{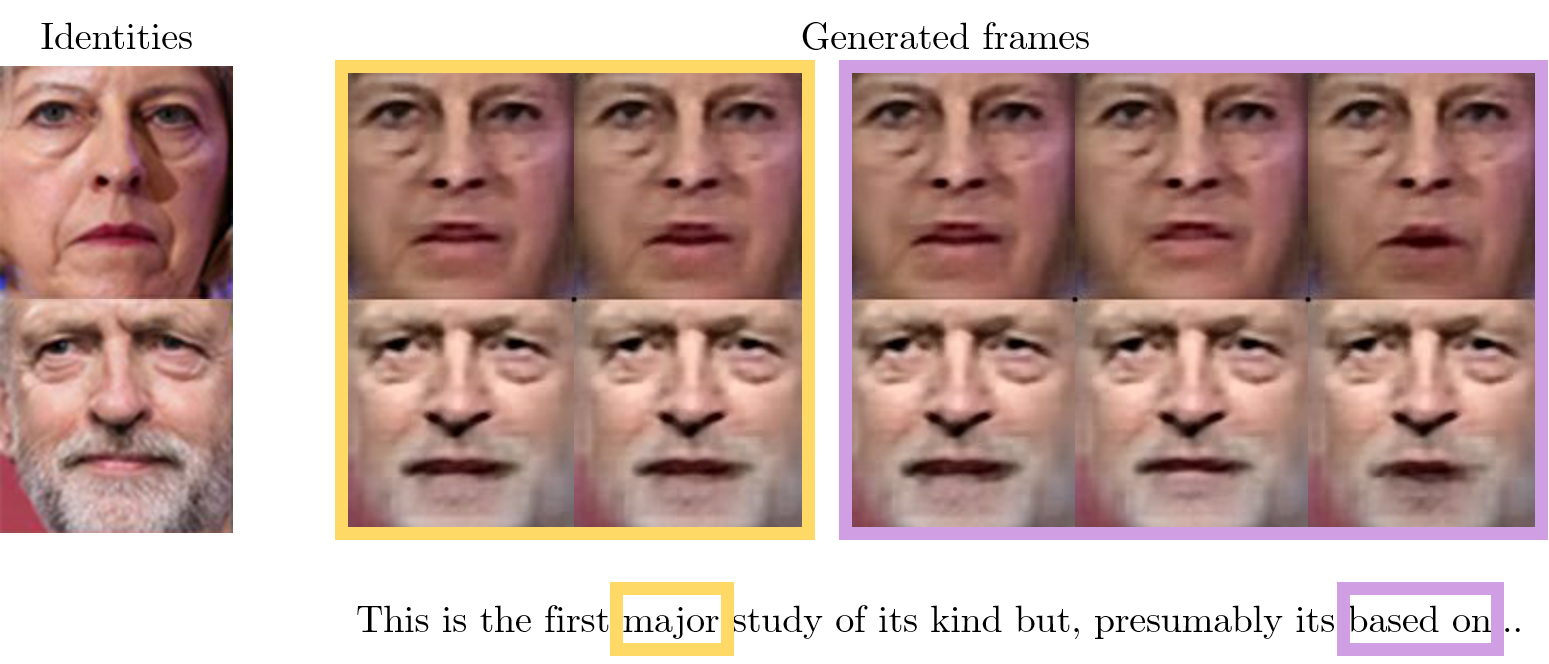}
	\caption{
		{\bf Top row:} 	Identity 1 and the corresponding generated frames;
		{\bf Middle row:} 	Identity 2 and the corresponding generated frames; 
		{\bf Bottom row:} 	Captions of the audio segment.
		\textbf{Best seen in video form.}
	}
	\label{fig:vis}
\end{figure}

Figure~\ref{fig:vis} shows a visualization of the output of the model (the frames of the two segments highlighted in the captions ``major'' and ``based on''). Note, the movement of the mouths of the two examples reflect the sound of each word not unlike phoneme-to-viseme correspondences.

\subsection{Preserving Identity with Skip Connections}
\label{subsec:skip}
\begin{figure}[h]
	\centering
	\includegraphics[width=0.24\textwidth]{images/faces/im1.jpg}~
	\includegraphics[width=0.24\textwidth]{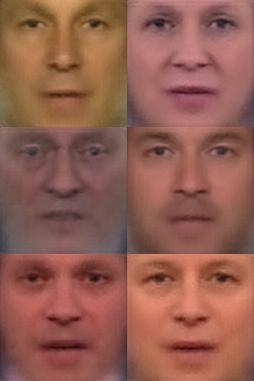}~
	\includegraphics[width=0.24\textwidth]{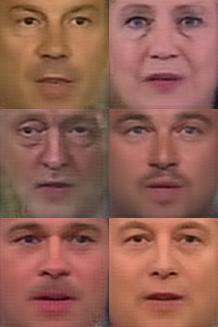}~
	\includegraphics[width=0.24\textwidth]{images/faces/im3.jpg}\\
	(a)~~~~~~~~~~~~~~~~~~~~~~~~~~~~~~~~~(b)~~~~~~~~~~~~~~~~~~~~~~~~~~~~~~~~~(c)~~~~~~~~~~~~~~~~~~~~~~~~~~~~~~~~~(d)
	\caption{
	{\bf (a)} 	Original still image to animate (input to the identity encoder);
	{\bf (b)} 	Output frames without skip connection; 
	{\bf (c)} 	Output frames with skip connection and one input image;
	{\bf (d)} 	Output frames with skip connection and five input images.
	It is clear that the skip connection helps to carry facial features
	over from the identity image to the generated video frames.}
	\label{fig:still}
\end{figure}

Figure~\ref{fig:still} shows a set of generated faces and various target identities (original stills). 
We observe that the skip connections are crucial to carry
facial features from the input image to the generated output
-- without these, the generated images lose defining facial
features of target identities, as shown in the middle column.
The skip connections at earlier layers (\eg after conv1)
were not used as it encouraged the output image to be too similar
to the still input, often restricting the mouth shapes that
we want to animate.

\subsection{Preserving Identity with Multiple Still Images}
\label{subsec:multieg}
Instead of a single image specifying the unique identity, five
distinct images are concatenated channel-wise resulting in an input dimension of
$112\times112\times15$,  and the {\em conv1} layer of the
identity encoder are modified to ingest inputs of these dimensions. 
In training these examples
are sampled with a similar strategy to that of
Figure~\ref{fig:samplingIdentity}, but multiple `identity' images are
sampled, instead of just one.

As can be seen in Figure~\ref{fig:still}, having multiple image
examples for the unique identity enhances the quality of the generated
faces. There are two reasons for this: first, with multiple example
images as input, it is likely that the network now has access to
images of the person showing the mouth open as well as closed.  Thus,
it has to hallucinate less in generation as, in principle, more can be
sourced directly from the input images; Second, although the faces are
aligned prior to the identity encoder, there are minor variations in
the movement of the face other than the lips that are not relevant to
the speech, from blinking and microexpression. The impact
of these minor variations when extracting unique identity features
is reduced by having multiple still images of
the same person.

\subsection{Application: Lip Transplant/Re-dubbing Videos}
\label{subsec:stitch}

The Speech2Vid model can be applied to 
visually re-dub a source video with a 
different segment of spoken audio.
The key stages are as follows:
(i) obtain still images from the source video for identity;
(ii) generate the face video for the given audio and identity
 using the Speech2Vid model;
(iii) re-align the landmarks of the generated video to the
source video frames, and
(iv) visually blend the aligned face with the source video frame.

\begin{figure}[h]
	\centering
	\includegraphics[width=0.48\textwidth]{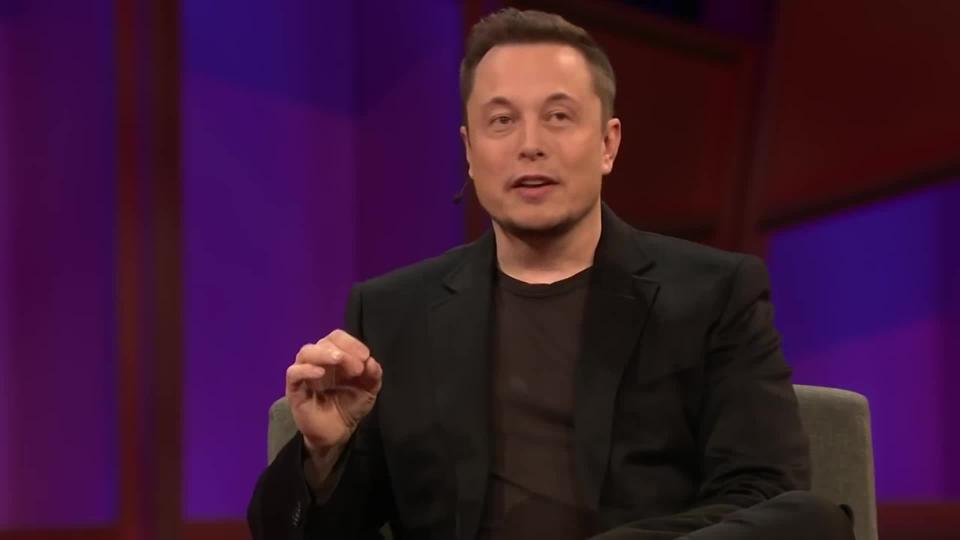}
	\includegraphics[width=0.48\textwidth]{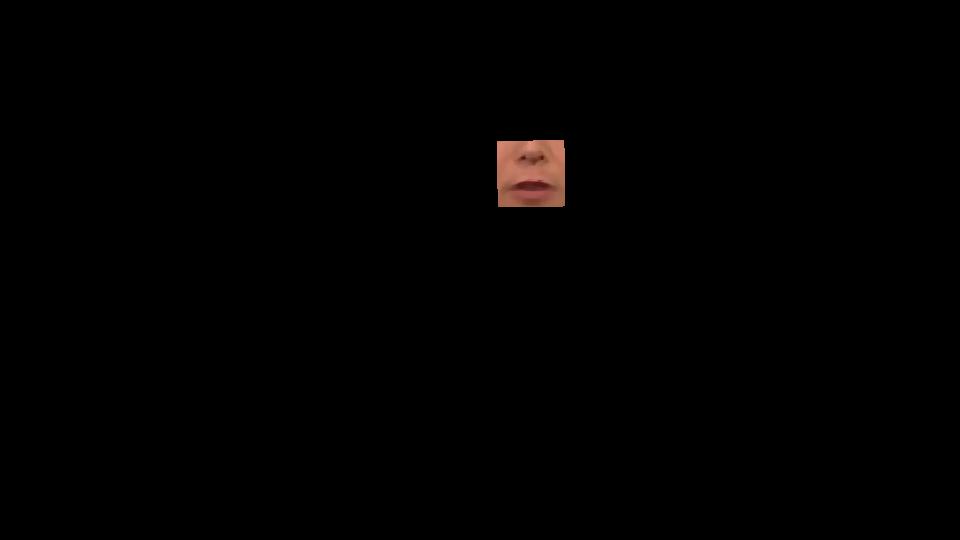}\\
	\vspace{1pt}
	\includegraphics[width=0.48\textwidth]{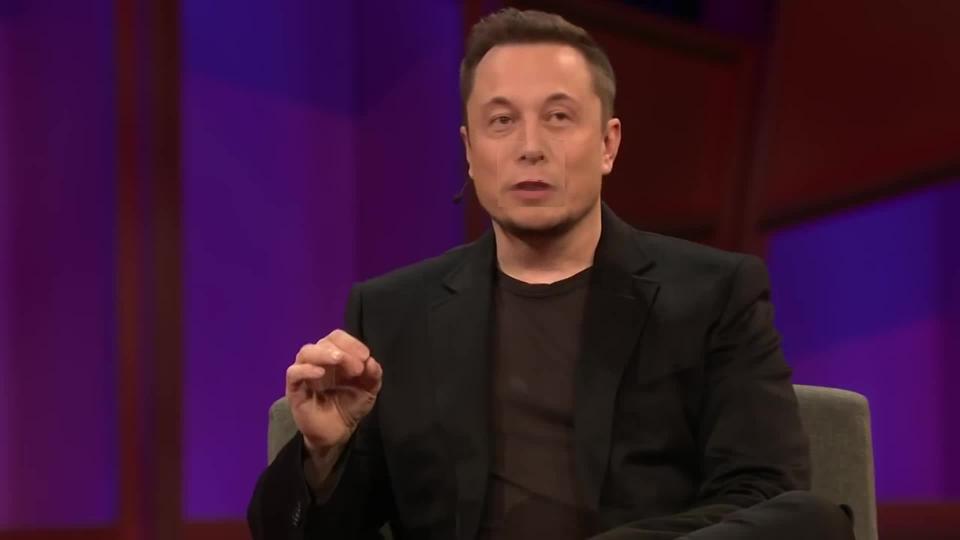}
	\includegraphics[width=0.48\textwidth]{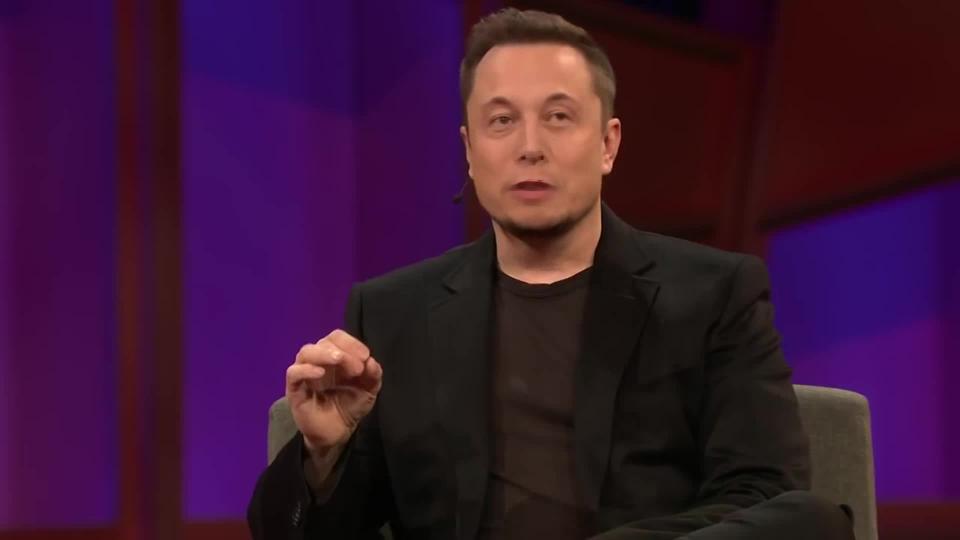}
	\caption{
	{\bf Top left:} 		Original still image;
	{\bf Top right:} 		Generated mouth region, aligned with the original (target) face; 
	{\bf Bottom left:} 		Generated mouth region, superimposed on the original face.
	{\bf Bottom right:} 		Generated mouth region, blended with the original face.}
	\label{fig:stitch}
\end{figure}

\subsub\noindent\textbf{Alignment.} 
Facial landmarks in the target video is
determined using the method of~\cite{Kazemi14}. 
A similarity transformation is used to align the 
generated face with the original face in the 
target image. Figure~\ref{fig:stitch} (right)
shows the generated face in alignment with 
the original face.

\subsub\noindent\textbf{Poisson editing.} 
The Poisson blending algorithm~\cite{Perez03} blends two images together by
matching gradients with boundary conditions.
We use this technique to match the generated face with the source video frame,
as shown in Figure~\ref{fig:stitch}.
This can be used to blend the face from the same, or different
identity to the source video frame. 

\subsub\noindent\textbf{Discussion.} 
This method can be used to blend the generated face as a whole,
or to match only the lower half of the face.
We qualitatively find that we strike the best balance between
image naturalness and movement naturalness by only blending
the lower half of the face, from just below the eyes.

\section{Summary and extensions}
\label{sec:conc}
We have demonstrated that the Speech2Vid model is able to generate
videos of any identity speaking from any source of input
audio. This work shows that there is promise in generating video
data straight from an audio source. We have also shown that re-dubbing
videos from a different audio source (independent of the original
speaker) is possible. 


One clear extension is to add a quantitative performance measure of our models. This is not a straightforward task as there is no definitive performance measure of generative models for a specific domain. In natural image generation, Salimans~\etal~\cite{salimans2016} proposed a scoring system dependent on an image's softmax output when fed into a network trained on a classification task \eg the inception network trained on ImageNet \cite{Szegedy15}. One possible option is to have a lip-specific inception score using networks trained on a lip-specific task~\cite{Chung17}.

Moving forward, this
model can be applied to computer facial animation relying only on
audio. 

\subsub\noindent\textbf{Acknowledgements.} Funding for this research is provided by the EPSRC Programme Grant Seebibyte EP/M013774/1. Amir Jamaludin is funded by the RCUK CDT in
Healthcare Innovation EP/G036861/1.
We would like to thank Aravindh Mahendran for helpful discussions. 

{\small
\bibliographystyle{splncs03}
\bibliography{shortstrings,vgg_local,vgg_other,mybib}

\begin{thebibliography}{10}
\providecommand{\url}[1]{\texttt{#1}}
\providecommand{\urlprefix}{URL }

\bibitem{charles2016virtual}
Charles, J., Magee, D., Hogg, D.: Virtual immortality: Reanimating characters
  from tv shows. In: Computer Vision--ECCV 2016 Workshops. pp. 879--886.
  Springer (2016)

\bibitem{Chatfield14}
Chatfield, K., Simonyan, K., Vedaldi, A., Zisserman, A.: Return of the devil in
  the details: Delving deep into convolutional nets. In: Proc. BMVC. (2014)

\bibitem{Chung17}
Chung, J.S., Senior, A., Vinyals, O., Zisserman, A.: Lip reading sentences in
  the wild. In: Proc. CVPR (2017)

\bibitem{Chung16}
Chung, J.S., Zisserman, A.: Lip reading in the wild. In: Proc. ACCV (2016)

\bibitem{Chung16a}
Chung, J.S., Zisserman, A.: Out of time: automated lip sync in the wild. In:
  Workshop on Multi-view Lip-reading, ACCV (2016)

\bibitem{fan2015photo}
Fan, B., Wang, L., Soong, F.K., Xie, L.: Photo-real talking head with deep
  bidirectional lstm. In: Acoustics, Speech and Signal Processing (ICASSP),
  2015 IEEE International Conference on. pp. 4884--4888. IEEE (2015)

\bibitem{garrido2015vdub}
Garrido, P., Valgaerts, L., Sarmadi, H., Steiner, I., Varanasi, K., P{\'e}rez,
  P., Theobalt, C.: Vdub: Modifying face video of actors for plausible visual
  alignment to a dubbed audio track. In: Computer Graphics Forum. vol.~34, pp.
  193--204. Wiley Online Library (2015)

\bibitem{GoodfellowPMXWOCB14}
Goodfellow, I.J., Pouget{-}Abadie, J., Mirza, M., Xu, B., Warde{-}Farley, D.,
  Ozair, S., Courville, A.C., Bengio, Y.: Generative adversarial nets. In:
  Advances in Neural Information Processing Systems 27: Annual Conference on
  Neural Information Processing Systems 2014, December 8-13 2014, Montreal,
  Quebec, Canada. pp. 2672--2680 (2014),
  \url{http://papers.nips.cc/paper/5423-generative-adversarial-nets}

\bibitem{HintonSalakhutdinov2006b}
Hinton, G.E., Salakhutdinov, R.R.: Reducing the dimensionality of data with
  neural networks. Science  313(5786),  504--507 (Jul 2006),
  \url{http://www.ncbi.nlm.nih.gov/sites/entrez?db=pubmed&uid=16873662&cmd=showdetailview&indexed=google}

\bibitem{pix2pix2016}
Isola, P., Zhu, J.Y., Zhou, T., Efros, A.A.: Image-to-image translation with
  conditional adversarial networks. arxiv  (2016)

\bibitem{karpathy2015deep}
Karpathy, A., Fei-Fei, L.: Deep visual-semantic alignments for generating image
  descriptions. In: Proc. CVPR. pp. 3128--3137 (2015)

\bibitem{Kazemi14}
Kazemi, V., Sullivan, J.: One millisecond face alignment with an ensemble of
  regression trees. In: Proceedings of the IEEE Conference on Computer Vision
  and Pattern Recognition. pp. 1867--1874 (2014)

\bibitem{Kim_2016_VDSR}
Kim, J., Lee, J.K., Lee, K.M.: Accurate image super-resolution using very deep
  convolutional networks. In: Proc. CVPR (June 2016)

\bibitem{King09}
King, D.E.: Dlib-ml: A machine learning toolkit. The Journal of Machine
  Learning Research  10,  1755--1758 (2009)

\bibitem{KingmaW13}
Kingma, D.P., Welling, M.: Auto-encoding variational bayes. CoRR  abs/1312.6114
  (2013), \url{http://arxiv.org/abs/1312.6114}

\bibitem{Krizhevsky12}
Krizhevsky, A., Sutskever, I., Hinton, G.E.: {ImageNet} classification with
  deep convolutional neural networks. In: NIPS. pp. 1106--1114 (2012)

\bibitem{Lienhart01}
Lienhart, R.: Reliable transition detection in videos: {A} survey and
  practitioner's guide. International Journal of Image and Graphics  (Aug 2001)

\bibitem{Lucas81}
Lucas, B.D., Kanade, T.: An iterative image registration technique with an
  application to stereo vision. In: Proc.~of the 7th International Joint
  Conference on Artificial Intelligence. pp. 674--679 (1981),
  \url{citeseer.nj.nec.com/lucas81optical.html}

\bibitem{Nagrani17}
Nagrani, A., Chung, J.S., Zisserman, A.: Voxceleb: a large-scale speaker
  identification dataset. In: INTERSPEECH (2017)

\bibitem{van2016conditional}
van~den Oord, A., Kalchbrenner, N., Espeholt, L., Vinyals, O., Graves, A.,
  et~al.: Conditional image generation with pixelcnn decoders. In: Advances in
  Neural Information Processing Systems. pp. 4790--4798 (2016)

\bibitem{OwensIMTAF16}
Owens, A., Isola, P., McDermott, J.H., Torralba, A., Adelson, E.H., Freeman,
  W.T.: Visually indicated sounds. In: CVPR. pp. 2405--2413. IEEE Computer
  Society (2016),
  \url{http://dblp.uni-trier.de/db/conf/cvpr/cvpr2016.html#OwensIMTAF16}

\bibitem{Parkhi15}
Parkhi, O.M., Vedaldi, A., Zisserman, A.: Deep face recognition. In: Proc.
  BMVC. (2015)

\bibitem{pathak2016context}
Pathak, D., Krahenbuhl, P., Donahue, J., Darrell, T., Efros, A.A.: Context
  encoders: Feature learning by inpainting. In: Proc. CVPR. pp. 2536--2544
  (2016)

\bibitem{Perez03}
Perez, P., Gangnet, M., Blake, A.: Poisson image editing. ACM Transactions on
  Graphics  22(3),  313--318 (2003)

\bibitem{ReedAYLSL16}
Reed, S.E., Akata, Z., Yan, X., Logeswaran, L., Schiele, B., Lee, H.:
  Generative adversarial text to image synthesis. In: Balcan, M.F., Weinberger,
  K.Q. (eds.) ICML. JMLR Workshop and Conference Proceedings, vol.~48, pp.
  1060--1069. JMLR.org (2016),
  \url{http://dblp.uni-trier.de/db/conf/icml/icml2016.html#ReedAYLSL16}

\bibitem{ronneberger2015u}
Ronneberger, O., Fischer, P., Brox, T.: U-net: Convolutional networks for
  biomedical image segmentation. In: International Conference on Medical Image
  Computing and Computer-Assisted Intervention. pp. 234--241. Springer (2015)

\bibitem{salimans2016}
Salimans, T., Goodfellow, I.J., Zaremba, W., Cheung, V., Radford, A., Chen, X.:
  Improved techniques for training gans. CoRR  abs/1606.03498 (2016),
  \url{http://arxiv.org/abs/1606.03498}

\bibitem{Szegedy15}
Szegedy, C., Liu, W., Jia, Y., Sermanet, P., Reed, S., Anguelov, D., Erhan, D.,
  Vanhoucke, V., Rabinovich, A.: Going deeper with convolutions. In: Proc. CVPR
  (2015)

\bibitem{Vedaldi15}
Vedaldi, A., Lenc, K.: Matconvnet: Convolutional neural networks for matlab.
  In: Proc. ACMM (2015)

\bibitem{Vinyals15}
Vinyals, O., Toshev, A., Bengio, S., Erhan, D.: Show and tell: A neural image
  caption generator. In: Proceedings of the IEEE Conference on Computer Vision
  and Pattern Recognition. pp. 3156--3164 (2015)

\bibitem{wan2013photo}
Wan, V., Anderson, R., Blokland, A., Braunschweiler, N., Chen, L., Kolluru, B.,
  Latorre, J., Maia, R., Stenger, B., Yanagisawa, K., et~al.: Photo-realistic
  expressive text to talking head synthesis. In: INTERSPEECH. pp. 2667--2669
  (2013)

\bibitem{Xu15}
Xu, K., Ba, J., Kiros, R., Courville, A., Salakhutdinov, R., Zemel, R., Bengio,
  Y.: Show, attend and tell: Neural image caption generation with visual
  attention. arXiv preprint arXiv:1502.03044  (2015)

\bibitem{zhang2016colorful}
Zhang, R., Isola, P., Efros, A.A.: Colorful image colorization. In: Proc. ECCV.
  pp. 649--666. Springer (2016)

\end{thebibliography}
}
\end{document}